\documentclass[final]{l4dc2026}

\usepackage{hyperref}
\usepackage{url}

\usepackage[utf8]{inputenc} 
\usepackage[T1]{fontenc}    
\usepackage{hyperref}       
\usepackage{url}            
\usepackage{booktabs}       
\usepackage{amsfonts}       
\usepackage{nicefrac}       
\usepackage{microtype}      
\usepackage{xcolor}         
\usepackage{tikz}
\usetikzlibrary{arrows.meta, positioning}
\usepackage{natbib}

\usepackage{algorithm2e}

\usepackage{floatflt}   
\usepackage{adjustbox}

\makeatletter\newcommand*{\IfPackageLoaded}{\@ifpackageloaded}\makeatother

\usepackage[T1]{fontenc}

\usepackage{amssymb}

\usepackage{bm}
\usepackage{dsfont}
\usepackage{newtxtext}
\usepackage{newtxmath}

\usepackage{algorithm}
\usepackage[noend]{algpseudocode}
\usepackage{bigstrut}
\usepackage{booktabs}
\usepackage{breakcites}
\usepackage{delimset}
\usepackage{float}
\usepackage[hybrid]{markdown}
\usepackage{microtype}
\usepackage{mleftright}\mleftright
\usepackage{multirow}
\usepackage{nicefrac}
\usepackage[normalem]{ulem}
\usepackage{wrapfig}
\usepackage[svgnames]{xcolor}
\usepackage{zref}

\newtoggle{include-notes}
\newtoggle{error-notes}
\toggletrue{include-notes}
\toggletrue{error-notes}

\newcommand*{\B}[1]{\ifmmode\bm{#1}\else\textbf{#1}\fi}
\newcommand*{\C}[1]{\mathcal{#1}}

\newcommand*{\Z}[1]{\mathds{#1}}

\DeclareMathOperator{\D}{\Z{D}}

\DeclareMathOperator{\E}{\Z{E}}

\newcommand*{\defeq}{\stackrel{\mathclap{\text{def}}}{=}}

\newcommand*{\eqn}[1]{\begin{align}#1\end{align}}

\newcommand*{\nn}{\nonumber}

\makeatletter
\def\fn{\gdef\@thefnmark{}\@footnotetext}
\makeatother

\makeatletter
\zref@newprop{mzheight}[0pt]{\the\ht\z@}
\zref@newprop{mzdepth}[0pt]{\the\dp\z@}
\newcount\c@@mz
\newcommand*{\the@mz}{mz\the\c@@mz}
\newcommand*{\@mz@list}{}    
\let\@mz@do\relax
\newcommand*{\mzreset}{%
  \begingroup
    \def\@mz@do##1{%
      \global\expandafter\let\csname mz@##1\endcsname\relax
    }%
    \@mz@list
    \global\let\@mz@list\@empty
  \endgroup
}
\newcommand*{\mzleft}[3]{%
  \@ifundefined{mz@#1}{%
    \global\advance\c@@mz\@ne
    \expandafter\xdef\csname mz@#1\endcsname{\the@mz}%
    \xdef\@mz@list{\@mz@list\@mz@do{#1}}%
  }{}%
  \expandafter\let\expandafter\@mz\csname mz@#1\endcsname
  \mleft#2%
  \expandafter\mathpalette\expandafter{%
    \expandafter\@mzleft\expandafter{\@mz}%
  }{#3}%
  \mright.\kern-\nulldelimiterspace
}
\newcommand*{\mzright}[3]{%
  \kern-\nulldelimiterspace
  \@ifundefined{mz@#1}{%
    \@latex@warning{Missing \string\mzleft{#1}}%
    \mleft.#2\mright#3%
  }{%
    \expandafter\let\expandafter\@mz\csname mz@#1\endcsname
    \mleft.%
    \expandafter\mathpalette\expandafter{%
      \expandafter\@mzright\expandafter{\@mz}%
    }{#2}%
    \mright#3%
  }%
}   
\newcommand*{\@mzleft}{%
  \@mzleftright lr%
}
\newcommand*{\@mzright}{%
  \@mzleftright rl%
}
\newcommand*{\@mzleftright}[5]{%
  \sbox0{$\m@th#4{}#5{}$}%
  \ifmeasuring@
  \else
    \begingroup
      \let\@auxout\@mainaux
      \zref@labelbyprops{#3#1}{mzheight,mzdepth}%
    \endgroup
  \fi
  \zifrefundefined{\@mz #2}{%
  }{%
    \dimen@=\zref@extract{#3#2}{mzheight}\relax
    \ifdim\dimen@>\ht0 %
      \ht0=\dimen@
    \fi
    \dimen@=\zref@extract{#3#2}{mzdepth}\relax
    \ifdim\dimen@>\dp0 %
      \dp0=\dimen@
    \fi
  }%
  \copy0\relax
}
\makeatother

\algnewcommand{\LeftComment}[1]{\Statex \(\triangleright\) #1}
\newcommand{\eqdotref}[1]{Eq.~(\ref{#1})}

\usepackage{mathtools}\mathtoolsset{mathic}

\title{Model-Based Reinforcement Learning \\ under Random Observation Delays}

\coltauthor{\Name{Armin Karamzade}, \Name{Kyungmin Kim}, \Name{JB Lanier}, \Name{Davide Corsi}, \Name{Roy Fox} \\
\addr Department of Computer Science\\
\addr University of California, Irvine\\
}

\begin{document}

\maketitle

\begin{abstract}

Delays frequently occur in real-world environments, yet standard reinforcement learning (RL) algorithms often assume instantaneous perception of the environment. We study random sensor delays in POMDPs, where observations may arrive out-of-sequence, a setting that has not been previously addressed in RL. We analyze the structure of such delays and demonstrate that naive approaches, such as stacking past observations, are insufficient for reliable performance. To address this, we propose a model-based filtering process that sequentially updates the belief state based on an incoming stream of observations. We then introduce a simple delay-aware framework that incorporates this idea into model-based RL, enabling agents to effectively handle random delays. Applying this framework to the Dreamer world-modeling scheme, our method consistently outperforms delay-aware baselines developed for MDPs and demonstrates robustness to delay distribution shifts during deployment. Additionally, we present experiments on simulated robotic tasks, comparing our method to common practical heuristics and emphasizing the importance of explicitly modeling observation delays.

\end{abstract}


\section{Introduction}

Despite the remarkable success of reinforcement learning (RL) across a wide range of domains, standard RL algorithms rely on the assumption of delay-free interaction with the environment. In practice, however, delays are pervasive and often unavoidable, particularly in real-world systems such as robotics, autonomous driving, and distributed control~\citep{abadia2021cerebellar, mahmood2018setting, fagundes2023communication}. These delays can occur at different stages of the pipeline, such as sensing, processing, and communication. They are generally divided into two types: (i) \textit{feedback delays}, the time lag in receiving observations, and (ii) \textit{execution delays}, describing the delay between selecting an action and its execution in the environment. While these are highly common in practical applications, they are typically ignored or oversimplified in the RL literature.

When delays are present, a common workaround in robotics is to issue “no-op” actions, effectively instructing the agent to wait until the delayed observation arrives~\citep{walsh2007planning}. However, this approach is often impractical or even unsafe. For instance, an autonomous vehicle detecting a sudden obstacle via a low-latency sensor cannot wait for other sensors, as delay may cause a collision. Even defaulting to braking may not be viable, and the vehicle must infer the safest evasive motion under the uncertainty caused by delayed perception.

Even when delays are explicitly considered, they are often addressed with simplifying assumptions that fail to capture their full complexity in real-world tasks. Existing approaches assume either a fully observable environment, as in Markov Decision Processes (MDPs)~\citep{katsikopoulos2003markov, liotet2021learning, derman2021acting, liotet2022delayed, wu2024variational}, or fixed delays in Partially Observable MDPs (POMDPs)~\citep{kim1987partially, karamzade2024reinforcement}. However, real-world systems often involve partial observability and random delays. Unlike MDPs, where each observation fully represents the state, POMDPs require integrating past observations to maintain, perhaps implicitly, a belief over it. With random delays, observations may arrive out-of-sequence (OOS), a phenomenon that does not arise in fixed-delay settings and is innocuous in MDPs, but critical in POMDPs, where relying only on the latest observation is insufficient for optimal control.

In this work, we consider random observation delays in POMDPs. To address the OOS phenomenon, we propose a latent-space filtering approach that enables effective learning in the presence of OOS observations imposed by random delays. By leveraging model-based approaches, our method forms a belief over the current latent state, given the set of available observations. In particular, the filtering process exploits a world model trained in the delayed environment to sequentially update the belief based on received observations. This belief state then serves as a sufficient statistic for policy learning under an incomplete set of observations to ensure actions are informed by only and all available inputs.

We evaluate on both synthetic control tasks and realistic simulated robotic environments. Our method outperforms existing approaches in fully observable MDP settings and is the only one capable of handling more realistic, partially observable scenarios. Notably, our approach also generalizes well to test-time delay distribution shifts: when trained on a wider delay distribution, it performs significantly better under shorter test-time delays and shows minimal performance degradation under longer ones. These results highlight its potential for real-world deployment, where delay patterns are often unknown in advance and nonstationary.

Here, we summarize the contributions of this paper as follows. (i) We study random observation delays in POMDPs and propose a framework that connects this setting to standard POMDP formulations. (ii) We introduce a filtering procedure for processing  OOS observations within model-based RL. (iii) We present the first method designed for this setting and describe how to integrate the filtering process into existing model-based RL algorithms. (iv) We conduct extensive experiments across diverse environments, demonstrating superior performance over baselines and strong generalization to unseen delay distributions.

\section{Preliminaries}\label{sec:preliminaries}

A \textit{Partially Observable Markov Decision Process (POMDP)} is a tuple $\C{M} = \langle S, A, \C{T}, r, \Omega, O, \gamma \rangle$, where $S$, $A$, and $\Omega$ denote the sets of states, actions, and observations, respectively. The transition dynamics is defined by $\C{T}(s' \mid s, a)$, the reward function by $r(s, a)$, and the observation (emission) probabilities by $O(o \mid s)$. At each timestep $t$, the environment is in state $s_t \in S$, the agent receives an observation $o_t \sim O(o_t \mid s_t)$, selects an action $a_t \in A$, receives reward $r_t = r(s_t, a_t)$, and transitions according to $s_{t+1} \sim \C{T}(s_{t+1} \mid s_t, a_t)$. The objective is to select actions that maximize the expected return $\E\left[\sum_{t=0}^\infty \gamma^t r_t\right]$, where $\gamma \in [0, 1)$ is the discount factor.

\subsection{Model-Based RL} \label{subsec:model-based-rl}

Recent model-based RL (MBRL) approaches focus on learning a latent dynamics model, or world model, that captures the environment's behavior and enables long-term prediction~\citep{ha2018recurrent, hansen2022temporal, micheli2022transformers, hafner2025mastering}. In this framework, the agent maintains a latent state $x_t$ governed by a parametrized transition model $p_\theta(x_t \mid x_{t-1}, a_{t-1})$, and generates observations through a decoder $p_\theta(o_t \mid x_t)$. Note that rewards are usually part of the model, but here we omit them for brevity. Since the latent state is not directly observable in training data, a variational posterior $q_\theta(x_{\le T} \mid o_{\le T}, a_{<T}) = \prod_t q_\theta(x_t \mid x_{t-1}, a_{t-1}, o_t)$, first proposed by~\citet{hafner2019learning}, can be used for the distribution of the latent state sequence of a particular observed episode $(o_{\le T}, a_{<T})$ of length $T$. The model is trained by maximizing an evidence lower bound~(ELBO) on the sequence log-likelihood, leading to the objective~\citep{hafner2019learning}:
\begin{align}
    \mathcal{L} = \sum_{t=0}^{T} \E_{q_\theta} \left[ \ln p_\theta(o_t \mid x_t) \right] 
- \E_{q_\theta} \left[ \D \left[ q_\theta(x_t \mid x_{t-1}, a_{t-1}, o_t) \,\Vert\, p_\theta(x_t \mid x_{t-1}, a_{t-1}) \right] \right], \label{eq:rssm-main}
\end{align}
where $\D$ is the Kullback–Leibler (KL) divergence. This objective encourages the latent state to retain sufficient information for reconstructing observations, while remaining consistent with the prior dynamics $p_\theta$. Throughout the text, we omit the dependence on $\theta$ whenever it is clear from context.

Many existing works exploit this or similar models for reinforcement learning and planning~\citep{ha2018recurrent,hafner2019learning, micheli2022transformers, zhang2023storm}. One notable example is Dreamer~\citep{hafner2025mastering}, which trains policies entirely within a learned Recurrent State Space Model (RSSM). Dreamer alternates between three key stages: (1) training the world model, (2) learning the policy through imagined trajectories, and (3) collecting new experiences.
\section{Random Observation Delay Environments}
\label{sec:delay_env}

\newcommand{\receivedatt}{\tilde{o}_t}
\newcommand{\receivedindicesbyt}{\mathcal{I}_t}
\newcommand{\receivedbyt}{o_{{\mathcal{I}_t}}}

We define a \textit{Random Observation Delay Environment} as a pair $\langle \mathcal{M}, \mathcal{D} \rangle$, where $\mathcal{M}$ is a standard POMDP and $\mathcal{D}$ is a distribution over non-negative integers representing stochastic delays. At each time step $t$, the observation generated by the environment is not revealed immediately, but is instead delivered after a random delay $d_t \sim \mathcal{D}$\footnote{Here we assume that delays are i.i.d for simplicity. However, our analysis holds for non-stationary or state-dependent delays with some adjustments.}. 
That is, $o_t$ becomes available at time $t + d_t$. The agent’s actual observation at time $t$ consists of all information scheduled to arrive at that step. We denote this~(possibly empty) set by $\receivedatt = \{ (o_\tau, \tau) : \tau + d_\tau = t \}$. We assume that the agent observes the timestamp of delivered observations, but not $d_\tau$ for undelivered observations. We further assume that all delays are bounded, with $D$ denoting the maximum possible delay.

In the fully observable setting, an equivalent delay-free MDP can be constructed by augmenting the state with the sequence of actions taken since the most recently observed state~\citep{katsikopoulos2003markov}. Similarly, in partially observable environments with constant delays, one can construct an equivalent delay-free POMDP by augmenting the state, while preserving the original observation space~\citep{karamzade2024reinforcement}. In both cases, the augmentation only needs to track a sequence of past actions. However, with random delays, this structure is no longer sufficient, and reducing a delayed environment to a standard POMDP becomes more complex.

\paragraph{Reducing to a standard POMDP.}

Under random observation delays, the agent may potentially receive any nonnegative number of observations at each timestep. The new observation space becomes $\widetilde{\Omega} = \bigcup_{k=0}^{D+1} \big( \Omega \times \Z{N} \big)^k$, where each new observation $\receivedatt$ is a set of observations delivered at time $t$, each paired with its original emission timestamp. To encode this within a standard POMDP, we augment the state with a buffer $u_t$ that stores all observations not previously received:
\[
u_t = \{ (o_\tau, \tau, d_\tau) : t - d_\tau \le \tau \le t \} \in U,
\]
where $U$ is the space of sets of observation–timestamp–delay tuples. The augmented state space is $\widetilde{S} = S \times U$, and the equivalent delay-free POMDP is given by the tuple $\langle \widetilde{S}, A, \widetilde{\C{T}}, \widetilde{r}, \widetilde{\Omega}, \widetilde{O}, \gamma \rangle$, with:
\begin{align*}
    \widetilde{r}\left((s_t, u_t), a_t\right) &= r(s_t, a_t), \\
    \widetilde{O}(\receivedatt \mid (s_t, u_t)) &= \delta\left(\receivedatt  = \{ (o_\tau, \tau) : (o_\tau, \tau, t - \tau) \in u_t \}\right), \\
    \widetilde{\C{T}}\left((s_{t+1}, u_{t+1}) \mid (s_t, u_t), a_t\right) &= \C{T}(s_{t+1} \mid s_t, a_t) \cdot \Z{P}_{\mathcal{D}}(u_{t+1} | u_{t}, s_{t+1}),
\end{align*}
where the observation function $\widetilde{O}$ deterministically returns all entries in the buffer that are scheduled for delivery at time $t$, and each new observation is drawn and stored in the buffer with its associated delay
$\Z{P}_\C{D}(u_{t+1} | u_t, s_{t+1}) = O(o_{t+1} | s_{t+1}) \C{D}(d_{t+1})$,
when
$u_{t+1} = \{(o_\tau, \tau, d_\tau) \in u_t: \tau + d_\tau > t\} \cup \{(o_{t+1}, t+1, d_{t+1})\}$
for any $o_{t+1}$ and $d_{t+1}$, and assigning probability 0 to any $u_{t+1}$ that does not have this form. 

This construction enables the use of standard POMDP algorithms, but at the cost of exponentially increased state and observation space sizes. In particular, the agent's belief must capture uncertainty not only over the latent state but also over pending, undelivered observations. This added complexity is fundamental to POMDPs with random delays, as observations may arrive out of sequence and cannot be ignored or replaced by the most recent one, unlike in MDPs.

\paragraph{World Model for the Reduced POMDPs.}

In the reduced POMDP formulation, the latent state at time $t$ must be inferred from the set of observations received up to that point. Therefore, $o_t$ would be replaced by $\tilde{o}_t$ in \eqdotref{eq:rssm-main} and the resulting ELBO becomes:
\begin{align}\label{eq:generic}
 \sum_{t=0}^{T} \E \left[ \ln p(\tilde{o}_t \mid x_t) \right] 
- \E \left[ \D \left( q(x_t \mid x_{t-1}, a_{t-1}, \tilde{o}_{t}) \,\Vert\, p(x_t \mid x_{t-1}, a_{t-1}) \right) \right].
\end{align}

While this formulation enables the use of standard MBRL tools in the delayed setting, it treats the set of received observations as a generic input and does not explicitly model the delay process. As a result, the model may learn unnecessarily complex dynamics to compensate for the partial and OOS nature of the observations, instead of exploiting the structure imposed by the delays.

\section{Delay-Aware Model-Based RL}
\label{sec:DAMBRL}

We introduce, in contrast to Section~\ref{sec:delay_env}, a structured latent-space filtering approach to address random delays and OOS inputs within the context of MBRL. By maintaining a belief over the current latent state using only received observations, the agent can act effectively under incomplete information. This section presents the belief update formulation and its integration into the MBRL framework.

\newcommand{\inbetweenindices}{\C{J}}
\newcommand{\kt}{\kappa_t}









\subsection{Out-of-Sequence Filtering via a World Model}

In the scheme of Section \ref{subsec:model-based-rl}, a standard model-based agent processes its observations as $q_\theta(x_t \mid x_{t-1}, a_{t-1}, o_t)$ and then bases its actions on $x_t$ as if it were the state of a fully observable MDP.
An agent operating under delayed observations, on the other hand, cannot reproduce the same latent process in time for the relevant decisions.
Instead, our agent can form a belief over the $x_t$ of an undelayed agent, and base its actions on this belief-state, as in a belief-state representation of a POMDP~\citep{kaelbling1998planning}.
In this case, the exact belief $\phi_t \coloneqq{} \Pr_q(x_t \mid \tilde{o}_{\le t}, a_{<t})$ is a sufficient statistic of the available information, namely the received observations $\tilde{o}_{\le t}$ and the past actions $a_{<t}$, for the latent state $x_t$.

It is worth highlighting two aspects of the temporal structure of this belief.
First, unlike standard belief-states, it may not be possible to compute $\phi_t$ fully sequentially only from $\phi_{t-1}$, $a_{t-1}$, and $\tilde{o}_t$.
The reason is that $\phi_{t-1}$ can lose information of past observations that only becomes informative once delayed inputs arrive.
For example, consider a state feature defined as the cumulative parity of a Bernoulli observation: if any observations are missing from $\tilde{o}_{\le t-1}$, then $\phi_{t-1}$ will have a uniform belief of their parity. This will prevent recovering that state feature in $\phi_t$ from $\phi_{t-1}$ and $\tilde{o}_t$, even if all remaining observations arrive by time $t$.
For sequential computation of $\phi_t$, we therefore consider the belief sequence $\phi_{\tau, t} \coloneqq{} \Pr_q(x_\tau | \tilde{o}_{\le t}, a_{<t})$, which may need partial recomputation at each step $t$.

The second temporal insight starts with noting that the “causal” variational structure of $q_\theta(x_{\le t} | o_{\le t}, a_{<t}) = \prod_\tau q_\theta(x_\tau | x_{\tau-1}, a_{\tau-1}, o_\tau)$ in Section \ref{subsec:model-based-rl} may be unable to represent the exact posterior of the generative process $p_\theta$, because in the latter $x_\tau$ generally depends on all observations $o_{\le t}$, rather than only on $o_{\le \tau}$ in $q_\theta$.
Despite creating a variational gap, in which the ELBO is generally not a tight lower bound, this structure of $q_\theta$ is justified by the need to use it online in deployment, as mentioned above.
A similar rationale reappears in the sequential computation of $\phi_{\tau, t}$, motivated here by greater sample efficiency and faster execution.


With these considerations in place, we define an auxiliary transition distribution $\psi_{\tau, t}$ over latent states that retroactively incorporates information available at time $t$ into the filtering process at time $\tau \le t$, using the learned models $q_\theta$ and $p_\theta$:
\[
\psi_{\tau, t}(x_\tau \mid x_{\tau-1}, a_{\tau-1}, [\tilde{o}_{\le t}]_\tau^\tau) =
\begin{cases}
q_\theta(x_\tau \mid x_{\tau-1}, a_{\tau-1}, o_\tau) & \text{if } (o_\tau, \tau) \in [\tilde{o}_{\le t}]_\tau^\tau, \\
p_\theta(x_\tau \mid x_{\tau-1}, a_{\tau-1}) & \text{otherwise},
\end{cases}
\]
where $[\tilde{o}_{\le t}]_{t_1}^{t_2}$ is the restriction of $\tilde{o}_{\le t}$ to timestamps in $[t_1, t_2]$, with $[\tau, \tau]$ representing a single timestamp. This auxiliary kernel $\psi$ serves as a time-dependent transition function that updates the state based on whether an observation at time $\tau$ becomes available by time $t$. It uses the variational posterior when $o_\tau$ is observed, and otherwise defaults to the prior dynamics model.
This lets us write an approximate $\phi_t = \phi_{t, t}$ recursively as
\eqn{
\phi_{\tau, t} &\defeq \Pr(x_\tau \mid \tilde{o}_{\le t}, a_{<t}) \nn \\
&= \E_{x_{\tau-1} | \tilde{o}_{\le t}, a_{<t}}[ \Pr(x_\tau \mid x_{\tau-1}, \tilde{o}_{\le t}, a_{<t}) ] \label{eq:phi_cond} \\
&= \E_{x_{\tau-1} | \tilde{o}_{\le t}, a_{<t}}[ \Pr(x_\tau \mid x_{\tau-1}, a_{\tau-1,\ldots,t-1}, [\tilde{o}_{\le t}]_\tau^t) ] \label{eq:phi_past} \\
&\approx \E_{x_{\tau-1} | \tilde{o}_{\le t}, a_{<t}}[ \Pr(x_\tau \mid x_{\tau-1}, a_{\tau-1}, [\tilde{o}_{\le t}]_\tau^\tau) ] \label{eq:phi_future} \\
&\approx \E_{(x_{\tau-1} | \tilde{o}_{\le t}, a_{<t}) \sim \phi_{\tau-1, t}}[ \psi_{\tau, t}(x_\tau \mid x_{\tau-1}, a_{\tau-1}, [{\tilde{o}_{\le t}}]_\tau^\tau) ], \label{eq:phi_approx}
}
where $\phi_{0, t}$ is the initial latent state distribution, \eqdotref{eq:phi_cond} follows from simple conditioning on $x_{\tau-1}$, \eqdotref{eq:phi_past} omits past observations and actions since they are separated by $x_{\tau-1}$, \eqdotref{eq:phi_future} omits future observations and actions as a “causal” variational approximation, and in \eqdotref{eq:phi_approx} the learned $q_\theta$ and $p_\theta$ approximate the corresponding distributions, and $\phi_{\tau-1, t}$ indicates that the same approximations are repeated in each step.

The belief-state we approximate here is a sufficient statistic for control with respect to the information actually available at time~\(t\) as no policy can exploit observations that have not yet arrived. Thus, conditioning the policy on \(\phi_t\) is optimal for that information set \citep{kaelbling1998planning}.
Note that, in the undelayed case, where $\tilde{o}_{\le t}$ includes all observations and $\psi$ always uses $q_\theta$, this process reduces to the standard latent state process.
More generally, we can represent an approximate belief distribution using particle filtering~\citep{ma2020discriminative} to capture uncertainty. Each step of the recursion in~\eqdotref{eq:phi_approx} is then represented by $K$ particles $\{x_{\tau}^k\}_{k=1}^K$ propagated independently through the model. 

\subsection{Incorporating into RL}

\definecolor{changes}{HTML}{1273db}

We present a general training procedure for incorporating belief inference into MBRL under delayed observations. This framework, outlined in Algorithm~\ref{alg:framework}, can be applied to any algorithm employing an RSSM-style world model. Modifications to the standard pipeline are highlighted in \textcolor{changes}{blue}. The key idea is to train the world model on complete, ordered trajectories as in undelayed settings, while training the policy on belief states inferred from partially observed sequences using \eqdotref{eq:phi_approx}.

\begin{algorithm2e}[t]
\caption{Delay-Aware MBRL}
\label{alg:framework}
\small
\KwIn{$\C{A}$: Model-Based RL algorithm optimizing objective~\eqref{eq:rssm-main}}
\tcp{Inference Mode}
\textcolor{changes}{Initialize $\kappa_{-1} = 0$ and $\hat{o}_{-1} = \emptyset$}\;
\For{time $t$ in episode}{
    Receive $\textcolor{changes}{\tilde{o}_t}$\; 
    \textcolor{changes}{Compute $\phi_t$ from $\phi_{\kappa_{t-1}-1, t-1}$ and $\hat{o}_{t-1} \cup \tilde{o}_t$ using \eqdotref{eq:phi_approx} with the model $(p, q)$}\;
    \textcolor{changes}{Find the time $\kappa_t$ of the first unreceived observation and checkpoint $\phi_{\kappa_t-1, t}$}\;
    \textcolor{changes}{Store $\hat{o}_t = [\hat{o}_{t-1} \cup \tilde{o}_t]_{\kappa_t}^t$, discard $[\hat{o}_{t-1} \cup \tilde{o}_t]_{\kappa_{t-1}}^{\kappa_t-1}$}\ that will no longer be needed;
    Execute action $a_t \sim \pi(\cdot \mid \textcolor{changes}{\phi_t})$\;
}
\tcp{Training Mode}
\ForEach{update step}{
    Collect data with $\pi$ using inference mode and store it in replay buffer $B$\;
    Use $\C{A}$ with sample $(o_{\le T}, a_{<T}, r_{\le T}, \textcolor{changes}{d_{<T}}) \sim B$ to update world model $(p, q)$\; 
    \textcolor{changes}{Compute beliefs $\phi_{<T}$ using \eqdotref{eq:phi_approx}}\;
    Use $\C{A}$ to update the policy $\pi(\cdot \mid \textcolor{changes}{\phi_t})$ (and critic $V(\textcolor{changes}{\phi_t})$, if applicable)\;
}
\end{algorithm2e}


During inference, the agent maintains a timestamped buffer of observations that may be needed for belief recomputation, when earlier observations eventually arrive. Let $\kappa_t$ be the earliest time of an observation unreceived by time $t$. Then $\phi_{\tau, t}$ for all $\tau < \kappa_t$ already uses all observations, and will thus never need recomputation. $\hat{o}_t \coloneqq{} [\tilde{o}_{\le t}]_{\kappa_t}^t$ is therefore a sufficient buffer for future recomputation. Importantly, $t - \kappa_t < d_{\kappa_t} \le D$, so this buffer has at most $D$ observations. In practice, one could set this maximum delay to some pre-defined constant that trades-off inference memory with performance. 

In each inference step $t$, the agent adds new observations $\tilde{o}_t$ to the buffer $\hat{o}_{t-1}$, recomputes $\phi_t$ from the latest checkpoint $\phi_{\kappa_{t-1}-1, t-1}$ using available observations $[\tilde{o}_{\le t}]_{\kappa_{t-1}}^t$ using \eqdotref{eq:phi_approx}, and stores $\hat{o}_t = [\tilde{o}_{\le t}]_{\kappa_t}^t$.
The agent then selects an action from the policy defined on the belief space $\pi(\,\cdot\mid\phi_t)$.

While training happens in a delayed environment, world model training remains identical to the standard setting. This is possible because learning and data collection processes are decoupled. In particular, after each episode terminates, we wait until all pending observations arrive before storing the ordered trajectory in the replay buffer. To make policy training consistent with deployment, i.e., delay-aware, the replay buffer is augmented with the delays $d_t$.  Replaying these indices allows us to reconstruct the same partial buffers, recompute beliefs, and provide them as inputs to the downstream policy learning algorithm.

\paragraph{Training in Imagination.}

MBRL algorithms that model prior dynamics train the policy purely on imagined trajectories to improve sample efficiency. In Algorithm~\ref{alg:framework}, we noted that policy learning is identical to the delay-free case, except that the policy input is now the belief $\phi_t$ inferred from received observations. In delayed environments, an optimal policy requires simulating the belief dynamics under the delay distribution. However, because policy training relies on imagined trajectories, ground-truth observations needed to model belief dynamics are unavailable during imagination. As a result, training in imagination corresponds to a scenario where the agent receives no further observations. Despite this distribution shift, our experiments in Section~\ref{sec:unseen} indicate that the policy still generalizes well to unseen delay distributions.

\section{Experiments}
\label{sec:exp}

\newcommand{\DCAC}{\texttt{DCAC}~}
\newcommand{\Addressing}{\texttt{Encoding}~}

\newcommand{\Extended}{\texttt{Stack-Dreamer}~}
\newcommand{\Latent}{\texttt{DA-Dreamer}~}
\newcommand{\Wait}{\texttt{Wait}~}
\newcommand{\Latest}{\texttt{Memoryless}~}

We evaluate our method through two sets of experiments. Section~\ref{exp:main} covers fully observable MuJoCo environments~\citep{todorov2012mujoco}, comparing against delayed-MDP baselines. As no prior work addresses our setting, we adopt MDP-based methods for fair comparison. Section~\ref{exp:secondary} evaluates our method on four Meta-World environments~\citep{yu2019meta} with visual inputs, which are inherently partially observable. In Meta-World, we compare against practical heuristics commonly adopted in the presence of delays, as naïve alternatives in the absence of existing delay-aware methods.

Our method\footnote{The code is available at \href{https://github.com/indylab/DA-Dreamer}{https://github.com/indylab/DA-Dreamer}.} builds on Dreamer-v3~\citep{hafner2025mastering}, an MBRL algorithm that learns an RSSM-based world model (Section~\ref{subsec:model-based-rl}), modified according to Algorithm~\ref{alg:framework}. Most experiments use a single particle ($K=1$) to approximate the belief, for computational efficiency, as we observed no significant performance differences across different values of $K$ (Appendix~\ref{app:particle}). Dreamer’s world model consists of both deterministic and stochastic paths, preserving information over long horizons, which likely contributes to making single particle sufficient. The version of Dreamer that treats the received observations $\tilde{o}_t$ as generic inputs, represented by stacked frames, is referred to as \Extended (\eqdotref{eq:generic}), and the version using the \textbf{D}elay-\textbf{A}ware Algorithm~\ref{alg:framework} as \texttt{DA-Dreamer}. Each experiment is repeated 
for $5$ random seeds.

\subsection{Main Comparison with Baselines} \label{exp:main}

For MuJoCo environments, we compare with \DCAC~\citep{bouteiller2020reinforcement} and the best-performing method of~\cite{wang2023addressing}, referred to as "detach Encoding" in their work and, for clarity, as \Addressing here. Throughout, we denote the uniform distribution by $\mathcal{U}\{a, b\}$ and rounded truncated Gaussian by $\mathcal{N}^+(\mu, \sigma^2)$, where we round each sample of a truncated Gaussian to the nearest integer.


\subsubsection{Results}

\renewcommand{\arraystretch}{1.2}
\begin{table*}[t]
\centering
\small
\setlength{\tabcolsep}{4pt}
\caption{Expected return of methods for different delay distributions across MuJoCo environments. Results represent
the mean and 95\% confidence interval. Bold values indicate statistical significance.}
\label{tab:main}
\begin{adjustbox}{width=\textwidth}
\begin{tabular}{c|l|c|c|c|c}
\specialrule{1pt}{0pt}{0pt}
Delay & Environment & \DCAC & \Addressing & \Extended & \Latent \\
\specialrule{1pt}{0pt}{0pt}

\multirow{6}{*}{$\mathcal{U}\{0, 10\}$}
& HalfCheetah-v4 & $3841.43 \pm 802.96$ & $4189.26 \pm 401.81$ & $1959.96 \pm 839.58$ & $\mathbf{4985.40 \pm 253.10}$ \\
& Hopper-v4 & ${2394.67 \pm 721.94}$ & ${2373.70 \pm 206.46}$ & ${2694.22 \pm 416.69}$ & ${2251.36 \pm 413.62}$ \\
& Humanoid-v4 & $1062.81 \pm 248.87$ & $552.21 \pm 31.89$ & $522.13 \pm 41.48$ & $\mathbf{1854.26 \pm 205.04}$ \\
& HumanoidStandup-v4 & $145293.09 \pm 4314.40$ & $113240.31 \pm 11024.15$ & $93154.06 \pm 17839.01$ & $\mathbf{220017.11 \pm 23671.63}$ \\
& Reacher-v4 & ${-6.36 \pm 0.55}$ & ${-6.79 \pm 0.30}$ & ${-6.81 \pm 0.23}$ & ${-6.37 \pm 0.13}$ \\
& Swimmer-v4 & $40.53 \pm 1.55$ & $121.56 \pm 23.92$ & $\mathbf{346.58 \pm 2.51}$ & $\mathbf{347.00 \pm 5.64}$ \\
\hline
\hline
\multirow{6}{*}{$\mathcal{U}\{0, 20\}$}
& HalfCheetah-v4 & $2144.86 \pm 526.30$ & $\mathbf{4242.77 \pm 213.47}$ & $1045.15 \pm 179.75$ & $2958.56 \pm 54.19$ \\
& Hopper-v4 & $6.48 \pm 0.83$ & $1707.41 \pm 147.40$ & $\mathbf{2981.87 \pm 275.14}$ & $1713.85 \pm 343.09$ \\
& Humanoid-v4 & $112.08 \pm 76.92$ & $545.76 \pm 17.19$ & $418.24 \pm 42.17$ & $\mathbf{855.97 \pm 95.77}$ \\
& HumanoidStandup-v4 & $139806.95 \pm 20078.74$ & $118456.86 \pm 7541.70$ & $101241.22 \pm 4591.26$ & $\mathbf{195611.38 \pm 14140.51}$ \\
& Reacher-v4 & ${-6.59 \pm 0.44}$ & ${-7.17 \pm 0.24}$ & ${-6.71 \pm 0.21}$ & ${-7.06 \pm 0.16}$ \\
& Swimmer-v4 & $34.19 \pm 3.38$ & $117.84 \pm 20.27$ & $\mathbf{348.32 \pm 3.21}$ & $\mathbf{349.60 \pm 3.53}$ \\

\specialrule{1pt}{0pt}{0pt}
\end{tabular}
\end{adjustbox}
\end{table*}

Table~\ref{tab:main} reports the expected returns of all methods under two uniform delay distributions. \Latent \\achieves better performance than other methods in more environments, despite lacking the prior knowledge that the environments are fully observable, by inheriting this property from the underlying world model. In contrast, \Extended performs well in simpler settings but fails to scale to environments with larger observation spaces, such as Humanoid and HumanoidStandup. This supports our earlier hypothesis that a generic PODMP reduction that treats delayed observations as generic inputs leads to an unnecessarily complex latent space and fails to exploit the structure of delays. \DCAC shows a sharp performance drop in Hopper and Humanoid under $\mathcal{U}\{0, 20\}$. These environments may terminate early due to unsafe joint configurations, exposing the limitations of augmentation-based methods under longer delays. \Addressing demonstrates more stable performance but quite consistently underperforms \Latent. While the baselines rely on the MDP assumption and are not burdened by integrating past information, \Latent consistently outperforms them.

\subsubsection{Evaluation on Unseen Delay Distribution}\label{sec:unseen}

In this section, we evaluate how well each method generalizes in deployment to delay distributions different from the one used during training. All methods are trained with delay distribution $\mathcal{U}\{0, 20\}$, and evaluated on three test distributions: $\mathcal{U}\{0, 10\}$ representing shorter delays, $\mathcal{U}\{10, 20\}$ for longer one, and $\mathcal{N}^+(10, 1)$ a narrow Gaussian around the training mean.
Note that policy values may change significantly under a delay distribution shift, even if the training distribution supports the test one.

\begin{figure}[t]
   \centering
   \includegraphics[width=0.5\textwidth]{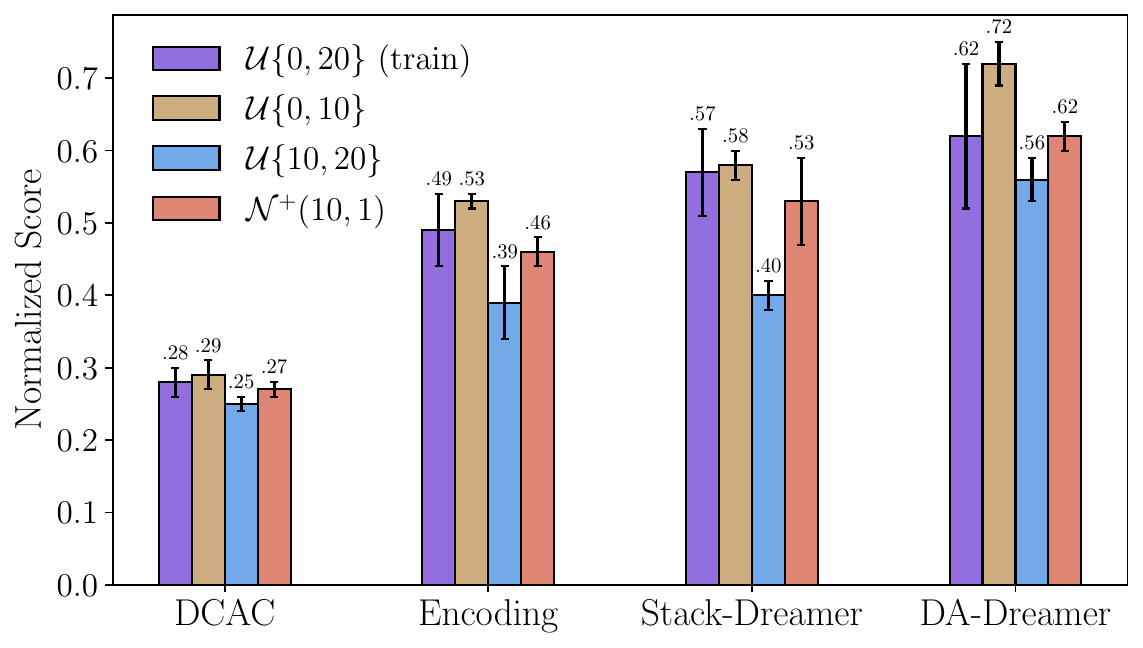}
   \caption{Normalized expected return under different test-time delay distributions. 
   Bars and caps represent the mean and 95\% confidence interval.}
   \label{fig:test_dist}
\end{figure}

Figure~\ref{fig:test_dist} reports the normalized expected return for each method, averaged across the same 6 environments. Normalization follows $\nu(R) = \frac{R - R_{\min}}{R_{\max} - R_{\min}}$, where $R_{\min}$ and $R_{\max}$ are the expected returns of a random policy and Dreamer trained without delay, respectively.  A well designed method for random delays should generally perform better under shorter delays. Otherwise, one could simply pad the delay to be longer.

Both \DCAC and \Extended show little improvement with shorter delays. \DCAC underperforms even within its training distribution, and its insensitivity to test-time shifts likely reflects more the weak overall performance rather than robustness. Its fixed-size state augmentation also prevents its applicability to delays longer than those seen during training. \Extended degrades sharply under longer delays, revealing the limits of treating observations as generic inputs without modeling temporal structure.
In contrast, \Latent generalizes well across all delay distributions, achieving much higher performance under shorter delays while remaining stable under longer ones. This is a desirable property, as delay distributions are often unknown or non-stationary in real-world settings, and methods trained on a distribution with wide support should ideally remain reliable at deployment. \Addressing exhibits a similar pattern to a lesser extent.

\subsection{Importance of Addressing Delays} \label{exp:secondary}

In Meta-World, because existing baselines are not applicable, we used two alternatives commonly employed in practice: (1) the \Latest agent uses the latest available observation in place of the current one; and (2) the \Wait method pauses to receive a new observation, where pausing is implemented by issuing no-op (zero) actions during waiting steps. While \Wait is not always feasible in practice, we include it for comparison purposes. To evaluate whether methods can complete the task \textit{in-time} while making delays more impactful, we reduce the default episode length in Meta-World environments to 50 steps, yet sufficient for the tasks considered.

\subsubsection{Results}

Table~\ref{tab:second} reports the final success rate, defined as the average proportion of successful episodes. The \Wait agent is only effective in the simplest task under short delays; as it fails entirely in most tasks, we exclude it from evaluations under longer delays. \Latest performs well under short delays but degrades quickly as delays increase, failing on more complex tasks. In contrast, \Latent consistently outperforms both baselines and maintains a high success rate even under long delays relative to the episode length. In stochastic environments, both \Latest and \Latent experience performance drops on harder tasks, which is expected given the increased task complexity when partial observations are delayed.

Additionally, we include experimental details and further results in the appendix, including stochastic environments, ablations on \Extended and \Latent, a study on the effect of number of the particles, visualizations of reconstructed observations, and training curves.

\renewcommand{\arraystretch}{1}
\begin{table*}[t]
\centering
\small
\setlength{\tabcolsep}{4pt}
\caption{Success rate of methods for different delay distributions in Meta-World environments. Results represent the mean and 95\% confidence interval.}
\label{tab:second}
\begin{tabular}{c|l|c|c|c}
\specialrule{.8pt}{0pt}{0pt}
Delay & Environment & \Wait & \Latest & \Latent \\
\specialrule{.8pt}{0pt}{0pt}

\multirow{4}{*}{$\mathcal{U}\{0, 5\}$}
& button-press-wall-v2 & $0.0 \pm 0.0$ & $\mathbf{0.99 \pm 0.01}$ & $\mathbf{0.90 \pm 0.09}$ \\
& drawer-close-v2 & $1.00 \pm 0.00$ & $1.00 \pm 0.00$ & $1.00 \pm 0.00$ \\
& plate-slide-v2 & $0.0 \pm 0.0$ & $0.69 \pm 0.22$ & $\mathbf{0.97 \pm 0.05}$ \\
& reach-v2 & $0.05 \pm 0.01$ & $\mathbf{1.00 \pm 0.00}$ & $\mathbf{1.00 \pm 0.00}$ \\
\hline
\hline
\multirow{4}{*}{$\mathcal{N}^+(10, 3)$}
& button-press-wall-v2 ($\alpha=0.25$)& - & $0.47 \pm 0.15$ & $0.49 \pm 0.26$ \\
& drawer-close-v2 ($\alpha=0.25$)& $0.14 \pm 0.04$ & $\mathbf{0.97 \pm 0.03}$ & $\mathbf{1.00 \pm 0.00}$ \\
& plate-slide-v2 ($\alpha=0.25$)& - & $0.39 \pm 0.13$ & $0.47 \pm 0.16$ \\
& reach-v2 ($\alpha=0.25$)& $0.0 \pm 0.0$ & $0.83 \pm 0.12$ & $\mathbf{1.00 \pm 0.00}$ \\
\hline
\hline
\multirow{4}{*}{$\mathcal{N}^+(20, 1)$}
& button-press-wall-v2 & - & $0.12 \pm 0.14$ & $\mathbf{0.91 \pm 0.13}$ \\
& drawer-close-v2 & - & $1.00 \pm 0.01$ & $1.00 \pm 0.00$ \\
& plate-slide-v2 & - & $0.05 \pm 0.06$ & $\mathbf{0.45 \pm 0.33}$ \\
& reach-v2 & - & $0.25 \pm 0.13$ & $\mathbf{1.00 \pm 0.00}$ \\

\specialrule{.8pt}{0pt}{0pt}
\end{tabular}
\end{table*}

\section{Related Work}\label{sec:related}

Research on delayed RL began with foundational works establishing unified frameworks for MDPs with observation and action delays~\citep{altman1992closed, katsikopoulos2003markov}. Early methods like dSARSA~\citep{schuitema2010control} used memoryless policies based on the last observation, but performance deteriorates quickly with increasing delays. Augmentation-based methods extend states using the last observation and subsequent actions; for instance, DCAC resamples trajectory fragments in hindsight~\citep{bouteiller2020reinforcement, haarnoja2018soft}, though such methods suffer from the curse of dimensionality. Model-based approaches infer the current state from extended states~\citep{walsh2007planning, derman2021acting} or learn compact belief representations~\citep{liotet2021learning}. Recent works explore design heuristics in deep RL~\citep{wang2023addressing} or use world models to predict the state~\citep{karamzade2024reinforcement, valensi2024tree}.
Other efforts apply imitation learning from undelayed experts~\citep{liotet2022delayed} or reformulate delayed RL as variational inference solved via behavior cloning~\citep{wu2024variational}, though these face policy mismatch. Auxiliary-task methods~\citep{wu2024boosting} introduce shorter delays to support training under long ones. BPQL~\citep{kim2023belief} avoids full augmentation by projecting critic evaluations onto the original state space but struggles under high stochasticity.
Most prior methods assume constant delays. Random delays remain underexplored, with few works~\citep{bouteiller2020reinforcement, wang2023addressing, valensi2024tree} tackling them—only in fully observable MDPs without OOS observations. In the partially observable setting, \cite{kim1987partially} studied lagged observations without proposing a learning method, while~\cite{karamzade2024reinforcement} addressed constant delays without OOS inputs. Our work fills this gap by targeting stochastic observation delays in POMDPs.


\section{Conclusion} \label{sec:conclusion}

We addressed random observation delays in POMDPs by proposing a model-based framework that effectively processes OOS observations for RL. Unlike prior methods, our approach does not assume full observability or fixed delays, making it applicable to more realistic scenarios. Experiments on synthetic and simulated robotic environments show that our method outperforms baselines even in MDP settings, remains effective under partial observability, and generalizes well to unseen delay distributions, an essential feature for real-world deployment.

A key limitation is the reliance on recursive filtering, which can accumulate one-step prediction errors and hinder scalability. Future work could explore more scalable architectures, such as Transformer-based models. Also, the filtering procedure incurs additional inference overhead that scales linearly with the maximum delay, increasing memory and update costs for large delays.

\section*{Acknowledgments}

Authors AK and KK were supported by Hasso Plattner Foundation Fellowship. Author DC was supported by DARPA ARC Award HR0011-24-3-0148. This work was supported by BSF Grant 2024079 and NSF Award 2321786.

\bibliography{ref}

\newpage
\appendix
\section{Experimental Details}

All methods and baselines were trained for $10^6$ environment steps. We used an action repeat of 2 for the Meta-World environments, resulting in $5 \times 10^5$ decision steps during training. For the baselines, we used the exact same hyperparameters as reported in their papers. For Dreamer-v3, we disabled the replay value loss, which prevents training the critic on data stored in the replay buffer; thus, the critic is only trained on generated trajectories during the policy learning phase. Additionally, we increased the number of classes in the discrete latent state representation to 32 and the number of neurons per layer to 512 for the Gym MuJoCo tasks only. In Meta-World, we used dense reward signals and $(64 \times 64)$ RGB images from \texttt{camera\_id=1} as observations.

\section{Additional Experiments}

\subsection{Stochastic environments}

While MuJoCo environments are deterministic apart from the initial state, we introduce added Gaussian noise with variance $\alpha$ into the normalized action space to evaluate the robustness of each method under stochastic environments. Table~\ref{tab:noisy} reports the performance of all methods under delay distribution $\mathcal{U}\{0, 10\}$ in the HalfCheetah environment across different noise levels. \Latent and \Addressing maintain competitive performance as noise increases, demonstrating greater robustness. In high-noise settings, \Latent achieves the best performance. \DCAC shows high performance in the noise-free setting for this environment (Table~\ref{tab:main}), but its performance degrades sharply under high stochasticity. Similarly, \Extended struggles even under mildly stochastic conditions. In such regimes, the agent must estimate a belief over latent states to act reliably. By explicitly computing this belief, \Latent effectively accounts for uncertainty, which is reflected in its strong performance.

\renewcommand{\arraystretch}{1.3}
\begin{table*}[htbp]
\centering
\small
\setlength{\tabcolsep}{4pt}
\caption{Exptected return under different levels of stochasticity of the environment($\alpha$) in HalfCheetah-v4. Results represent the mean and 95\% confidence interval.}
\label{tab:noisy}
\begin{adjustbox}{width=\textwidth}
\begin{tabular}{c|l|c|c|c|c}
\specialrule{1pt}{0pt}{0pt}
Delay & Environment & \DCAC & \Addressing & \Extended & \Latent \\
\specialrule{1pt}{0pt}{0pt}

\multirow{4}{*}{$\mathcal{U}\{0, 10\}$}
& HalfCheetah-v4 ($\alpha=0.2$)& $2061.48 \pm 726.16$ & $\mathbf{3538.54 \pm 244.46}$ & $1608.72 \pm 385.88$ & $\mathbf{3399.84 \pm 149.52}$ \\
& HalfCheetah-v4 ($\alpha=0.4$)& $1735.57 \pm 110.66$ & $\mathbf{2598.16 \pm 130.39}$ & $1252.71 \pm 406.18$ & $2347.83 \pm 58.20$ \\
& HalfCheetah-v4 ($\alpha=0.6$)& $1177.43 \pm 394.54$ & $\mathbf{1768.95 \pm 186.91}$ & $720.30 \pm 247.73$ & $\mathbf{1707.84 \pm 48.65}$ \\
& HalfCheetah-v4 ($\alpha=0.8$)& $623.95 \pm 303.60$ & $855.34 \pm 45.21$ & $336.37 \pm 146.56$ & $\mathbf{1093.30 \pm 80.53}$ \\
& HalfCheetah-v4 ($\alpha=1$)& $147.47 \pm 218.97$ & $383.75 \pm 110.70$ & $65.02 \pm 24.13$ & $\mathbf{587.99 \pm 60.00}$ \\
\specialrule{1pt}{0pt}{0pt}
\end{tabular}
\end{adjustbox}
\end{table*}

\subsection{Delay Aware Inference without Training}

\begin{figure}[h]
    \centering
    
    \subfigure[]{\includegraphics[width=0.75\linewidth]{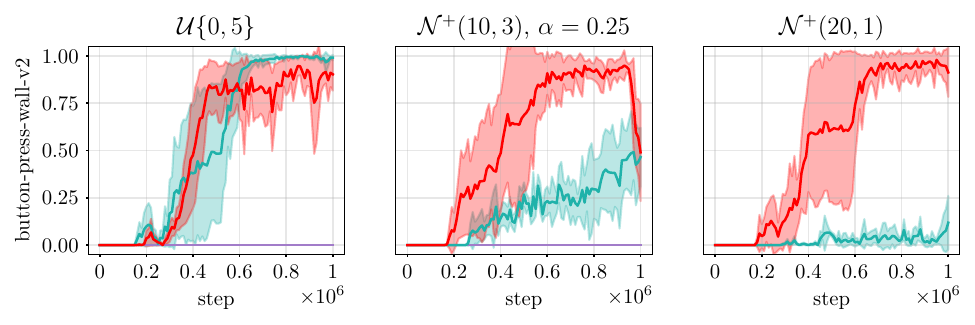}}
    
    \subfigure[]{\includegraphics[width=0.75\linewidth]{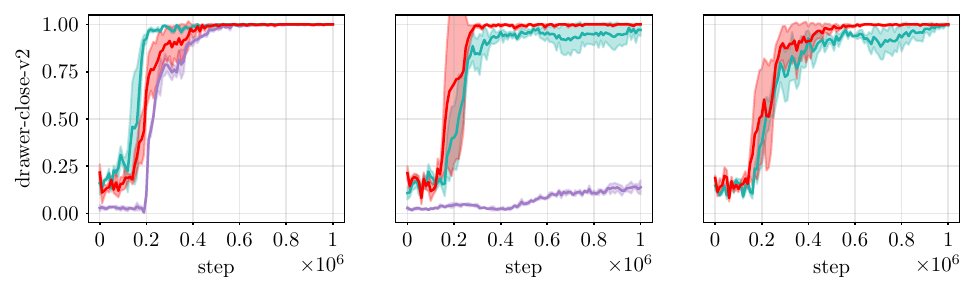}}
    
    \subfigure[]{\includegraphics[width=0.75\linewidth]{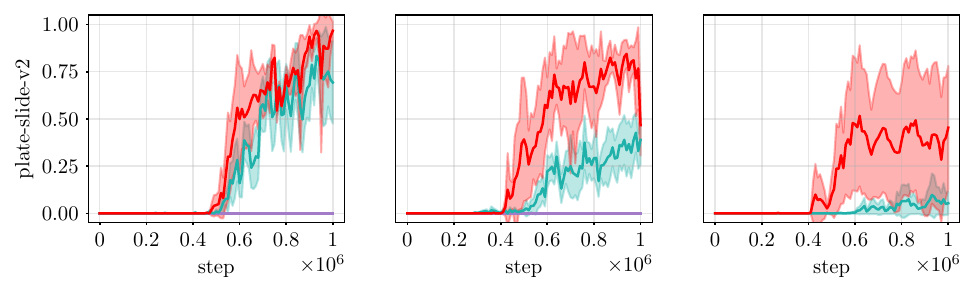}}
    
    \subfigure[]{\includegraphics[width=0.75\linewidth]{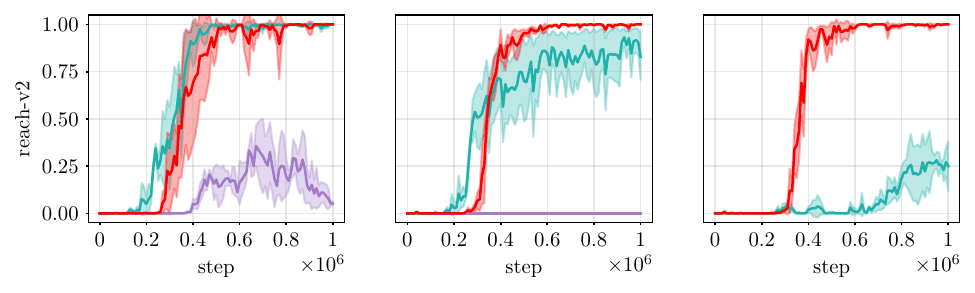}}
    
    \par\medskip
    
    \includegraphics[width=.65\linewidth]{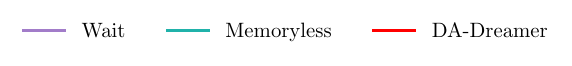}
  
  \caption{Learning curve of the methods on selected Meta-World environments.}
  \label{fig:training-gym}
\end{figure}

Table~\ref{tbl:agnostic} shows the final expected return of \texttt{DA-Dreamer} when trained in a standard (non-delayed) environment but deployed in a delayed environment. Compared to training directly in the delayed setting, it underperforms in half of the environments, shows similar performance in two, and achieves higher scores in HumanoidStandup. We speculate, based on Figure~\ref{fig:training-gym}, that the original method has not yet converged in HumanoidStandup, and with additional training, the performance gap may close. Overall, this approach appears promising for scenarios where the delay distribution is unknown during training, or when the agent must be trained once and deployed under varying, unknown delays.

\renewcommand{\arraystretch}{1.2}
\begin{table*}[htbp]
\centering
\small
\setlength{\tabcolsep}{4pt}
\caption{Expected return of the \texttt{DA-Dreamer} ablation on Gym environments. An asterisk (*) indicates similar performance. Results represent the mean and 95\% confidence interval.}
\label{tbl:agnostic}
\centering
\begin{tabular}{c|l|c}
\specialrule{.8pt}{0pt}{0pt}
Delay & Environment & \texttt{DA-Dreamer} (inference only) \\
\specialrule{.8pt}{0pt}{0pt}

\multirow{6}{*}{$\mathcal{U}\{0, 10\}$}
& HalfCheetah-v4 & $2642.02 \pm 644.15$ \\
& Hopper-v4 & $1692.25 \pm 1603.35$ \\
& Humanoid-v4 & $1808.95 \pm 631.46$ \\
& HumanoidStandup-v4 & $\mathbf{267659.35 \pm 52512.41}$ \\
& Reacher-v4 & $-6.13 \pm 0.23^*$ \\
& Swimmer-v4 & $349.99 \pm 2.53^*$ \\
\hline
\hline
\multirow{6}{*}{$\mathcal{U}\{0, 20\}$}
& HalfCheetah-v4 & $1384.41 \pm 233.52$ \\
& Hopper-v4 & $1430.39 \pm 1695.49$ \\
& Humanoid-v4 & $722.37 \pm 264.09$ \\
& HumanoidStandup-v4 & $\mathbf{207792.76 \pm 25154.57}$ \\
& Reacher-v4 & $-7.08 \pm 0.28^*$ \\
& Swimmer-v4 & $350.35 \pm 3.43^*$ \\
\specialrule{.8pt}{0pt}{0pt}
\end{tabular}
\label{tab}
\end{table*}

\subsection{Stacking Observations}

\begin{figure}[h]
    \centering
    \subfigure[]{\includegraphics[width=0.32\linewidth]{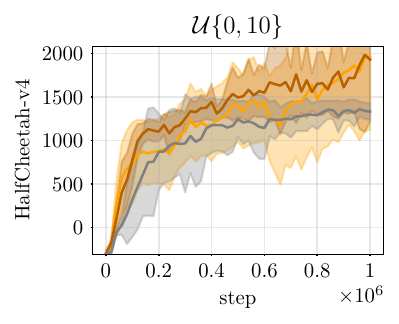}}
    \subfigure[]{\includegraphics[width=0.32\linewidth]{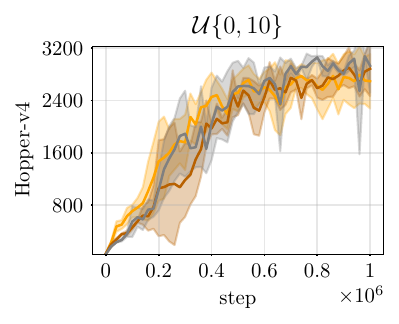}}
    \subfigure[]{\includegraphics[width=0.32\linewidth]{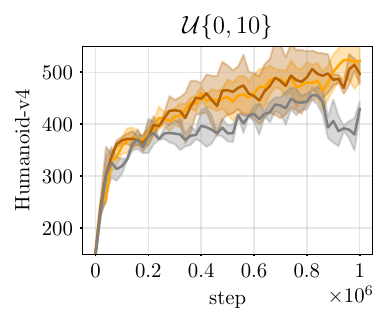}}
    
    \subfigure[]{\includegraphics[width=0.32\linewidth]{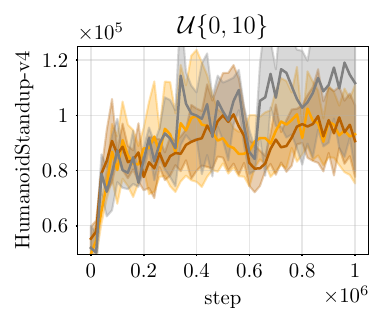}}
    \subfigure[]{\includegraphics[width=0.32\linewidth]{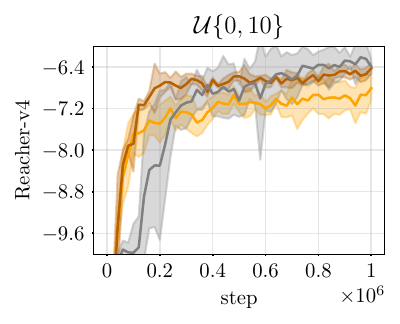}}
    \subfigure[]{\includegraphics[width=0.32\linewidth]{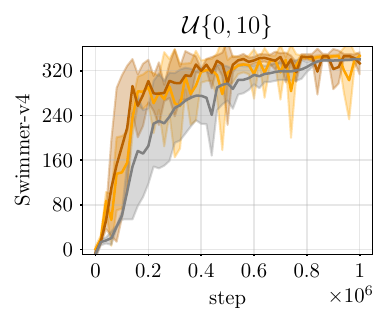}}

    \includegraphics[width=.8\linewidth]{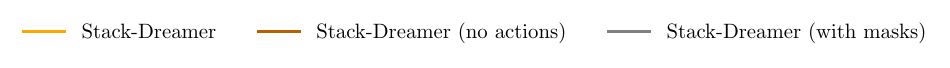}  
  
    \caption{Learning curve of \texttt{Stack-Dreamer} variants.}
    \label{fig:different-extended}
\end{figure}

In Figure~\ref{fig:different-extended}, we experiment with different ways of stacking delayed information in \texttt{Stack-Dreamer}. The default version, reported in the main text, inputs the previous $D$ observations and actions, with missing observations filled with zeros. We also evaluate a variant that removes actions from the input, as they are already represented in the latent state from the previous time step, and another variant that additionally includes a mask indicating which observations have been received. As shown, all variants perform similarly, with no significant differences observed between them.

\subsection{Number of Particles}
\label{app:particle}

\begin{figure}[h]
    \centering
    
    \subfigure[]{\includegraphics[width=0.39\linewidth]{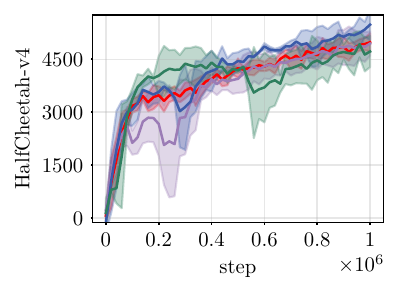}}
    \subfigure[]{\includegraphics[width=0.39\linewidth]{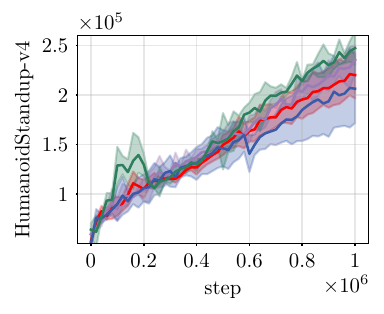}}
  
    \includegraphics[width=.5\linewidth]{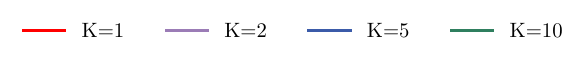}

    \caption{Performance vs different number of particles.}
    \label{fig:particles}
\end{figure}

Figure~\ref{fig:particles} shows the performance of \texttt{DA-Dreamer} with different numbers of particles $K$. In our implementation, we simply concatenate the $K$ particles, though various alternatives exist for combining them~\citep{ma2020particle}. As shown, increasing the number of particles generally improves performance, but the gains are not substantial. We hypothesize that this is because the Dreamer-v3 world model includes both deterministic and stochastic components, with the deterministic part capable of retaining information along a trajectory. To balance performance with computational cost, we used $K = 1$ in the main experiments.

\subsection{Visualizing Reconstructed Frames}

\begin{figure}[h]
    \centering

    \subfigure[$\mathcal{N}^+(10, 3)$ and $\alpha=0.25$]{\includegraphics[width=\linewidth]{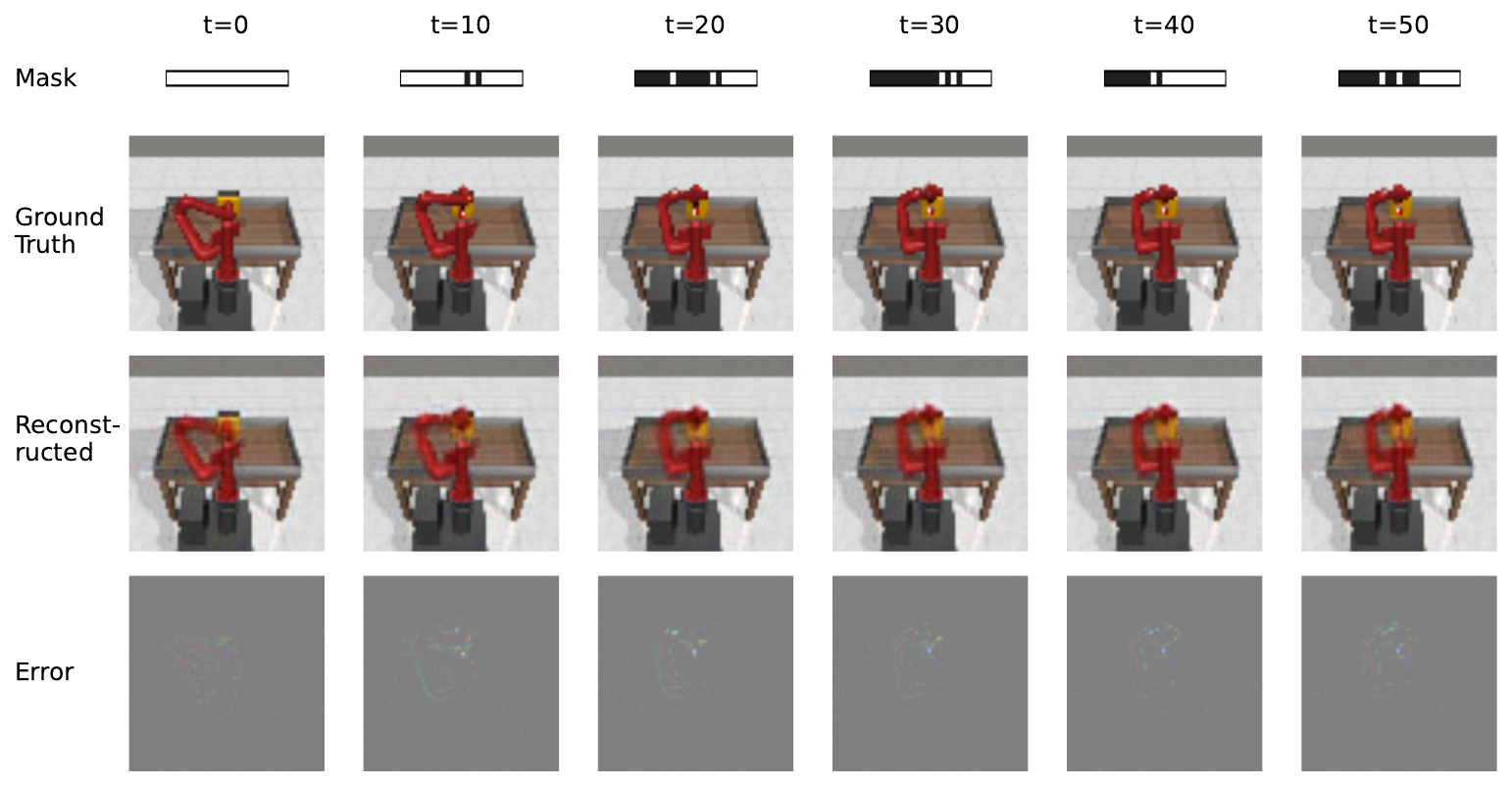}}
    \hspace{2\linewidth} 

    \subfigure[$\mathcal{N}^+(20, 1)$]{\includegraphics[width=\linewidth]{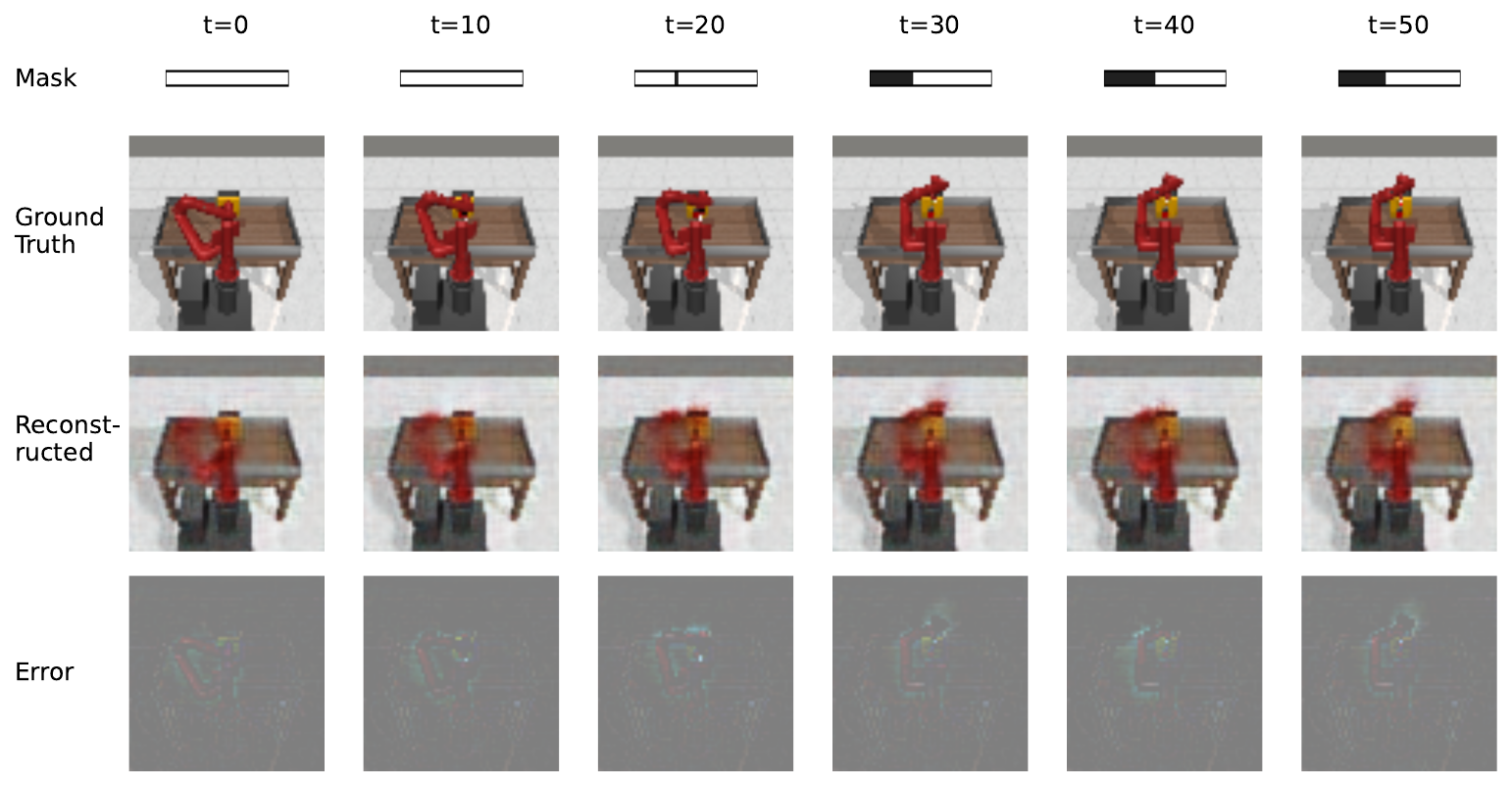}}
  
    \caption{Reconstructing the current observation in delayed button-press-v2 from the computed belief state. The mask indicates the arrival of past observations (black cells denote received observations), with the rightmost cell representing the current timestep.}
    \label{fig:recon}
\end{figure}

Figure~\ref{fig:recon} shows reconstructed observations from delayed information in \texttt{DA-Dreamer}. As seen, for shorter delays (a), the reconstructed frames are sharper compared to longer delays (b). This is expected, as longer delays increase uncertainty in the belief state about the current environment state. Consequently, the belief assigns more weight to nearby states, resulting in blurrier reconstructed frames, as depicted.

\clearpage

\section{Training Curves}

Below, we include the training plots for the experiments presented in Table~\ref{tab:main} and Table~\ref{tab:second}, respectively. Shaded regions represent 95\% confidence interval.

\subsection{Gym MuJoCo}

\begin{figure}[h]
    \centering
    
    \subfigure[]{\includegraphics[width=.49\linewidth]{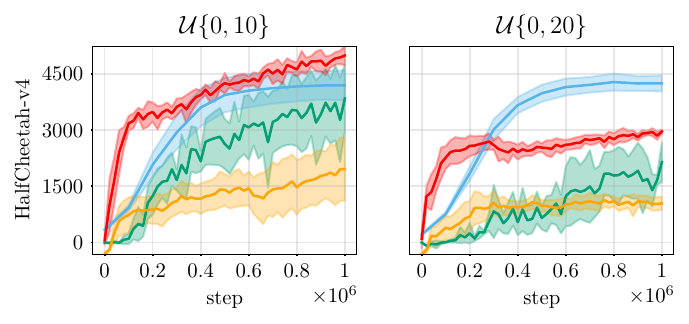}}\hfill
    \subfigure[]{\includegraphics[width=.49\linewidth]{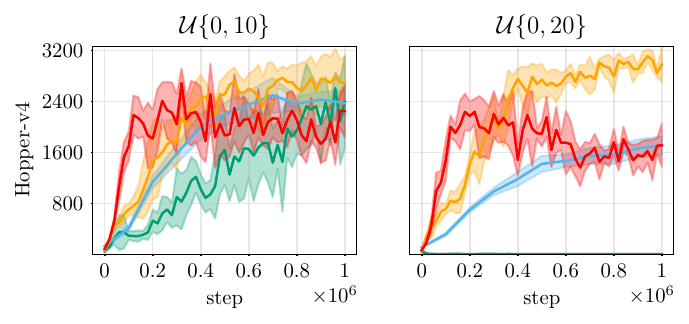}}
    
    \subfigure[]{\includegraphics[width=.49\linewidth]{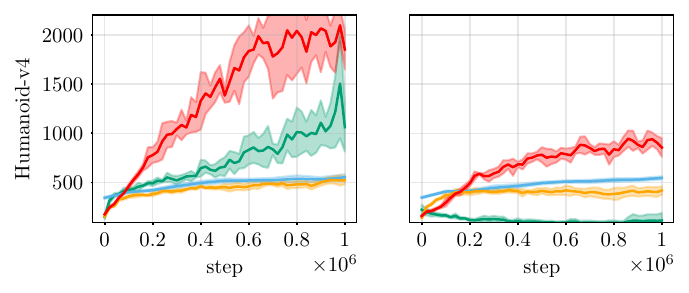}}\hfill
    \subfigure[]{\includegraphics[width=.49\linewidth]{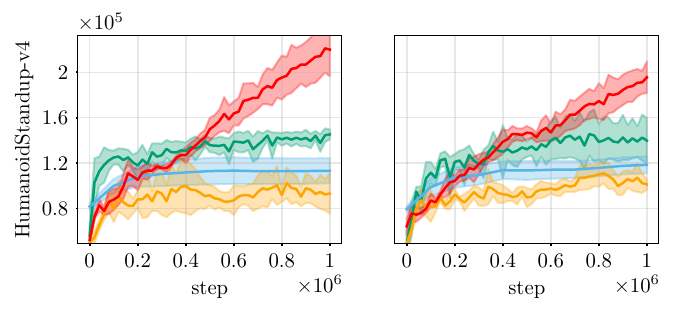}}
    
    \subfigure[]{\includegraphics[width=.49\linewidth]{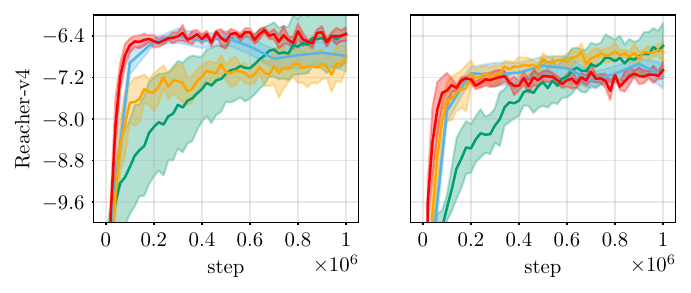}}\hfill
    \subfigure[]{\includegraphics[width=.49\linewidth]{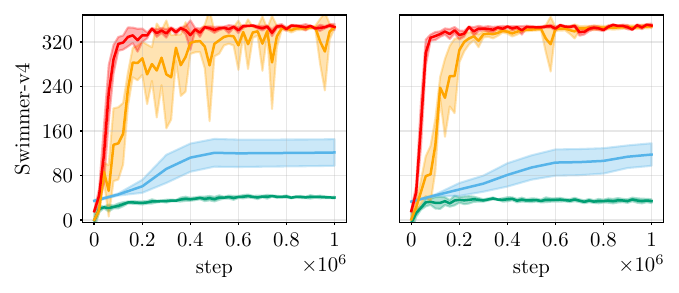}}
    
    \par\medskip
    
    \includegraphics[width=.75\linewidth]{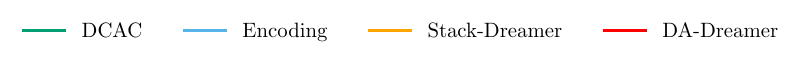}

    \caption{Learning curve of the methods on selected Gym environments.}
\end{figure}

\clearpage

\subsection{Meta-World}

\end{document}